\pdfoutput=1

\documentclass[11pt]{article}

\usepackage{emnlp2021}

\usepackage{times}
\usepackage{latexsym}

\usepackage[T1]{fontenc}

\usepackage[utf8]{inputenc}

\usepackage{microtype}

\newcommand{\pf}[1]{#1} 
\newcommand{\riz}[1]{#1} 

\usepackage{paralist}

\usepackage{graphicx}
\usepackage{amsmath,amssymb}
\usepackage{makecell}
\usepackage{longtable}
\usepackage{hhline}
\usepackage{multirow}
\usepackage[inline]{enumitem}
\DeclareMathOperator{\EX}{\mathbb{E}}
\usepackage[export]{adjustbox}

\title{What Makes a Scientific Paper be Accepted for Publication?}

\author{Panagiotis Fytas, Georgios Rizos, Lucia Specia \\
         Imperial College London \\ 
         \texttt{ \{pf2418, georgios.rizos12, l.specia\}@imperial.ac.uk}}
         
\begin{document}
\maketitle
\begin{abstract}

Despite peer-reviewing being an essential component of academia since the 1600s, it has repeatedly received criticisms for lack of transparency and consistency. We posit that recent work in machine learning and explainable AI provide tools that enable insights into the decisions from a given peer review process. 
We start by extracting global explanations in the form of linguistic features that affect the acceptance of a scientific paper for publication on an open peer-review dataset. 
Second, since such global explanations do not justify causal interpretations, we provide a methodology for detecting confounding effects in natural language in order to generate causal explanations,\pf{ under assumptions}, in the form of lexicons. 
Our proposed linguistic explanation methodology indicates the following on a case dataset of ICLR submissions: \begin{enumerate*}[label={\alph*)}]
\item  the organising committee follows, for the most part, the recommendations of reviewers, and,
\item the paper's main characteristics that led to reviewers recommending acceptance for publication are
originality, clarity and substance.
\end{enumerate*}
\end{abstract}

\section{Introduction}

The peer review process has been instrumental in academia in determining which papers meet the quality standards for publication in scientific journals and conferences. 
However, it has received several criticisms,
including inconsistencies among
review texts, review scores, and the final acceptance decision \citep{kravitz2010editorial},
arbitrariness between different reviewer groups \citep{Langford2015TheAO} as well as reviewer bias in ``single-blind'' peer reviews \citep{Tomkins12708}. 

Explainable AI (XAI) techniques have been shown to provide global understanding of the data behind a model.\footnote{Among the several definitions for explainability and interpretability \cite{def1, rudin2018stop, Murdoch_2019}, we follow the definition of explainability as a \textit{plausible} justification for the prediction of a model \citep{rudin2018stop}, and use the terms interpretability and explainability interchangeably.} 
Assume we build a classifier that predicts whether a paper is accepted for publication based on a peer review. Interpreting the decision of such a classifier can help comprehend what the reviewers value the most
on a research paper.
XAI allows us to elicit knowledge from the model about the data \cite{molnar2019}. However, in order to determine what aspects of a paper lead to its
acceptance, we need to interpret the peer review classifier explanations \textit{causally}. Generally, XAI methods do not provide such guarantees since most Machine Learning models simply detect correlations \cite{molnar2019}.
Nevertheless, in recent years there has been a trend towards enabling Machine Learning to adjust for causal inference \cite{schlkopf2019causality}.

Simply interpreting a non-causal classifier does not suffice when attempting to gain insight into human decision-making in the peer review process. For instance, words such as ``gan'', and ``efficiency'' appear to be important
in such a classifier (Section \ref{sec:data_expl}). We argue that correlation between such words and paper acceptance is confounded on the subject of the paper:
different subjects have different probabilities of acceptance, as well as 
different degrees to which ``efficiency'' is an important factor.
Since paper subject ground truth is not necessarily available, we can treat the abstract of a paper as a proxy for its subject, similar to \citet{CausalBERT}.

Previous approaches that aim to reconcile NLP with causal inference are limited either to binary treatment variables \cite{CausalBERT} or nominal confounders \cite{Pryzant2017PredictingSF, pryzant-etal-2018-interpretable, pryzant-etal-2018-deconfounded}. 
The aim of this paper is to
make a first consideration into using learnt natural language embeddings both as the treatment and the confounder.
We achieve this by generating deconfounded lexicons \citep{pryzant-etal-2018-interpretable} through interpreting causal models that predict the acceptance of a scientific paper,
based on their
peer reviews and confounded on their abstract. Our {\bf main contributions} are:


 \begin{itemize}
     \item We provide a methodology for developing text classifiers that are able to adjust for confounding effects located in natural language. Our evaluation is quantitative, reporting the Informativeness Coefficient measure \cite{pryzant-etal-2018-interpretable, pryzant-etal-2018-deconfounded}, and we also showcase the highest scoring words from the lexicons.
     \item We extend the classifier of \citet{pryzant-etal-2018-deconfounded} to use a black-box, instead of an interpretable classifier, which can be explained through model-agnostic tools, such as LIME \citep{lime}.
    \item We utilise our methodology to extract insights about the peer review process,
     given that certain assumptions hold. Those insights validate our perceptions of how the peer-review process works, indicating that the method makes meaningful reasonings.
    \item We develop novel models for the task of peer review classification, 
    which achieve a $15.79\%$ absolute increase in accuracy over previous state-of-the-art 
    \citep{ghosal-etal-2019-deepsentipeer}.

 \end{itemize}

In experiments with a dataset from ICLR 2017, we found that the organising committee mostly followed the recommendations of the reviewers, and that the reviewers focused on aspects such as originality, clarity, impact, and soundness of a paper, when suggesting whether or not the paper should be accepted for publication.


The paper is organised as follows: Section \ref{sec:related_word} presents related work regarding computational analyses of the peer-review process and explainability in NLP. Section \ref{sec:dataset} gives an overview of PeerRead, the peer review dataset we used in this paper. Section \ref{sec:data_expl} presents our initial exploration explanation of peer-reviewing, while Section \ref{sec:causal} describes our methodology to account for causality by generating deconfounded lexicons.

\section{Related Work}
\label{sec:related_word}

\textbf{Peer Review Analysis.} The PeerRead dataset \citep{PeerRead} is the only openly available peer review dataset. It contains submission data for arXiv and the NeurIPS, ICLR, ACL, and CoNLL conferences,
however, review data for both accepted and rejected submissions exist only for ICLR. The authors devised a series of NLP prediction baselines on the tasks of paper acceptance based on engineered features, as well as reviewer score prediction based on modelling abstract and review texts. Improvements on these baselines have been proposed in \citep{ghosal-etal-2019-deepsentipeer,milam}, by using better text representations as well as joint abstract/review modelling for predicting paper acceptance.  \citet{interspeechpaper} revisited the latter task on a dataset containing all the Interspeech 2019  conference abstracts and reviews (which is unfortunately not available) by utilising a review text fusion mechanism. The aforementioned studies model correlations between textual indices and the desired targets, without attempts towards explainability or identification of causal relationships. \citet{AMPERE},
utilise argumentation mining on review texts from the ICLR and UAI conferences. Whereas their methodology focuses on proposition type statistics and transitions, it is a distinct approach to understanding peer-reviewing compared to ours, which is based on XAI.


\textbf{Interpretability in NLP.} 
Interpreting aspects of the input as conducive to the model prediction aims to:
\begin{inparaenum}[a)]
\item increase model trustworthiness, and
\item allow for greater understanding of the data \citep{molnar2019}.
\end{inparaenum}
One way to do this is via \textbf{local explanations}, which explain the result of a single prediction, such as the view of high, learnt, \riz{class-agnostic} attention weights \citep{Bahdanau2015NeuralMT} applied to Recurrent Neural Network (RNN) hidden states, as proxies of word importance.
However, the use of attention weights as explanations has been controversial, as \citet{jain2019attention} have pointed out that for the same prediction, there can be counterfactual attentional explanations. Inversely, \citet{wiegreffe2019attentionnotnot} offer the justification that there may indeed exist multiple plausible explanations, out of which the attention mechanism captures one. \pf{An additional limitation of the attention mechanism is that the attention weights do not provide any class-specific insights, due to the use of the softmax activation function.}
Alternatively, the predictions of black-box classifiers can be interpreted through model-agnostic frameworks, such as LIME \citep{lime} and SHAP \citep{SHAP}. 

We are
interested in \textbf{global explanations}, which explain either a set of predictions or provide insights about the general behaviour of the model. Current research is limited either to feature-based classifiers \citep{10.1093/bioinformatics/btq134}, image classifiers \citep{kim2017interpretability} or combining local explanations to generate global explanations \citep{xaitree, globalxainn}.

\begin{figure}[]
    \centering
    \includegraphics[scale=0.35]{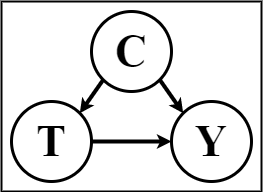}
    \caption{Graphical model depicting the confounding effect: C -- Confounder, T -- Treatment, Y -- Outcome.}
    \label{fig:confoundingpr}
\end{figure}

\textbf{Causal Inference in NLP.} Let us assume two observed random variables: the treatment $T$ and the outcome $Y$. Correlation between those two variables does not necessarily entail causation. Instead, this correlation could be the product of the confounding effect: a third random variable $C$ causes both of the observed random variables \cite{PetersJonas2017EoCI}. A depiction of the confounding effect can be observed in Figure \ref{fig:confoundingpr}. 
\pf{ \citet{fong-grimmer-2016-discovery} explore extracting treatments from text corpora and estimating their causal effect on human decisions. However, they require human annotators for the constructions of training and test sets.}
\citet{CausalBERT} provide a framework for treating text as a confounder, but they are limited to binary treatment variables. 
\citet{Pryzant2017PredictingSF, pryzant-etal-2018-interpretable, pryzant-etal-2018-deconfounded}  examine the problem of identifying a deconfounded lexicon (a set of linguistic features such as words or n-grams) from the text, which acts as a treatment variable. A deconfounded lexicon is ``predictive of a target variable'' but ``uncorrelated to a set of confounding variables'' \citep{pryzant-etal-2018-deconfounded}. However, their work is limited to using nominal confounders. In this paper, we explore the use of natural language both as the treatment variable and the confounder.

\section{Dataset}
\label{sec:dataset}

The PeerRead dataset \citep{PeerRead} consists of 14.7k papers and 10.7k peer reviews, including the meta-reviews by the editor.
The papers are from different venues, having been collected with different methods, which leads to a non-uniform dataset.
For instance, arXiv papers do not contain any reviews, and consequently, they are 
out of scope for our study.
Furthermore, the NeurIPS section contains 2k reviews for only accepted papers. Therefore, using them will lead to a significant imbalance in our data. For those reasons, we decided to examine the 1.3k ICLR 2017 reviews of PeerRead. The ICLR 2017 section is divided into training, validation and test sets with an $80\%$-$10\%$-$10\%$ split. In Table \ref{tab:peerread_descr} of the Appendix, we can observe the proportion of the accepted to rejected paper in the various partitions of ICLR 2017.

PeerRead suffers from various data quality issues, which we have resolved. Firstly, we have removed several empty and duplicate reviews. More importantly, for ICLR 2017 the meta-reviews which contain the review and final decision of the conference chair are not marked as \textit{meta-reviews} (the meta-review boolean field is marked as false) as they should have been \cite{PeerRead}. Instead, they are all marked as normal reviews with the title ``ICLR Committee Final Decision''. We explicitly treat them as meta-reviews, a differentiation that we believe is crucial in an XAI study like ours.
Notably, subsequent studies that utilise PeerRead, like DeepSentiPeer \cite{ghosal-etal-2019-deepsentipeer},
do not mention whether they have addressed this issue. This hinders a direct comparison with results from DeepSentiPeer as, expectedly, according to our experiments, the use of the meta-reviews significantly increases the performance of classifiers. 

\section{Data Exploration with Global XAI}
\label{sec:data_expl}


We make an initial exploration into global explanations in the form of lexicons by mapping each word to an aggregate of scores corresponding to local explanations of PeerRead test sample predictions, as suggested
both by the creators of SHAP \citep{xaitree} and by \citet{pryzant-etal-2018-deconfounded}. We follow a train-test split setup in our modelling experiments, and only extract explanations from the test set, following \citet{molnar2020pitfalls}.

We experimented with three different classification tasks:
\begin{inparaenum}[a)]
    \item the \textit{Final Decision Peer Review Classifier} fuses the multiple peer reviews to predict the final acceptance decision for a paper, 
    \item the \textit{Meta-Review Classifier} uses the singular meta-review to predict the final acceptance decision for a paper, and
    \item the \textit{Individual Peer Review Classifier} predicts the recommendation of the reviewer for a single review, where we consider as accepted the papers with scores above 5, given a range of 1-10.
\end{inparaenum}

\subsection{Interpreting Two Classifier Architectures}


After preliminary experiments with Transformer Language Models (TLMs), we have opted to use the \texttt{SciBERT\textsubscript{SciVocab} uncased} model \citep{SciBERT} trained on 1.7M papers from Semantic Scholar \citep{SemanticScholar} that contain a total of 3.17B tokens for text representation.

We experimented with two different text modelling approaches, each allowing for a different local explanation generation technique. In our \textbf{first approach}, we used the [CLS] token of our \texttt{SciBERT} model for \textit{review text representation}. We truncate the end of reviews longer than $512$ words. The effect of this truncation is not extremely adverse since only $10.5\%$ of the reviews exceed the maximum length. More details about the length of peer reviews are included in the Appendix. We are not fine-tuning our model due to the risk of overfitting on our small dataset of $1.3k$ reviews. For the Meta Review and the Individual Review Classifiers, an additional feed forward neural layer is required, followed by a sigmoid activation function to predict the decisions. In the case of the Final Decision Peer Review Classifier, multiple reviews exist, leading to an equal number of \texttt{SciBERT} representations. In order to avoid introducing an arbitrary ordering among the reviews, we fused them into a single representation using an attention mechanism, following \citet{interspeechpaper}. We used two more attention layers to produce variant fused embeddings, and concatenated them into a single vector of fixed dimensionality in order to simulate multi-headed attention \cite{attentionisallyouneed}.

For the \textbf{second approach}, we only explore the Meta-Review and Individual Peer Review Classifiers. We treat each review as a sequence of word representations given by \texttt{SciBERT}, which we further process using an RNN layer with a Gated Recurrent Unit cell (GRU), followed by an attention mechanism, which is used to provide interpretability to our model. The output of the RNN layer is a vector produced through attention pooling. This vector is input to a feed-forward layer to produce the classification decision.




For the \texttt{SciBERT} [CLS] approach, we treat our model as a black-box and use \textbf{LIME} \citep{lime} to produce explanations for each test set sample. More important words will have larger absolute scores. 
For the \texttt{GRU}-based approach, the attention mechanism is used to generate local explanations. More important words will have larger weights. Global explanation scores are then computed as the average of either the LIME score or the attention weights. The implementation details of our model are discussed in the Appendix.

\subsection{Results}

{\footnotesize
\begin{table*}
\centering
{\fontsize{9.5pt}{9.5pt}\selectfont
\begin{tabular}{llccc}
\hline
\textbf{Model Type} & \textbf{Model} & \textbf{Accuracy} & \textbf{Macro-F1} & \textbf{Weighted-F1} \\
\hline
\multirow{3}{*}{Baselines}                                                    & Majority Baseline                           & 59.52                     & 0.3731                     & 0.4442                        \\
                                                                              & PeerRead (only paper)                       & 65.30                      & N/A                         & N/A                            \\
                                                                              & DeepSentiPeer (paper \& review)             & 71.05                     & N/A                & N/A                   \\ \hline
\multirow{2}{*}{\begin{tabular}[c]{@{}l@{}}  Meta-Review \end{tabular}} & SciBERT                                     & 89.47                     & 0.8899                     & 0.8947                        \\
                                                                              & GRU                                         & 86.84                     & 0.8661                     & 0.8698                        \\ \hline
\begin{tabular}[c]{@{}l@{}}Final Decision \\ Peer Review\end{tabular} & SciBERT                                     & 86.84                     & 0.8606                     & 0.8676                        \\ \hline
\end{tabular}
}
\caption{Performance of the classifiers that predict the final decision prediction (i.e., whether a paper is accepted for publication on ICLR 2017) for a paper on the test set of PeerRead. We see that using Meta-reviews in training yields an advantage, as does peer review fusion.}
\label{tab:final_baseline}
\end{table*}

}

{\footnotesize
\begin{table*}
\centering
{\fontsize{9.5pt}{9.5pt}\selectfont
\begin{tabular}{llccc}
\hline                                                             \textbf{Model Type}          &\textbf{ Model} & \textbf{Accuracy} & \textbf{Macro-F1} & \textbf{Weighted-F1}
\\ \hline
Baselines                                                                                  & Majority Baseline      & 50.76                     & 0.3367                     & 0.3418                        \\ \hline
\multirow{2}{*}{\begin{tabular}[c]{@{}c@{}}Individual Peer Review\end{tabular}} & \multicolumn{1}{l}{SciBERT}                & 80.30                     & 0.8026                     & 0.8028                        \\ 
                                                                                           & \multicolumn{1}{l}{GRU}                    & 70.45                     & 0.7007                     & 0.7012                        \\  \hline
\end{tabular}
}
\caption{Performance on the Individual Peer Review Prediction (i.e., whether a reviewer suggests that a paper is accepted for publication), using on the test set of PeerRead.}
\label{tab:indiv_baseline}
\end{table*}
}

The validity of the explanations in drawing conclusions about peer reviewing is tied to the model predictive performance \citep{molnar2020pitfalls}. Table \ref{tab:final_baseline} summarises the hold-out performance on the PeerRead test set 
for the tasks related to predicting the final acceptance recommendation. Although not directly comparable, as explained in Section \ref{sec:dataset}, we also report the baseline performances of PeerRead \citep{PeerRead} and DeepSentiPeer  \citep{ghosal-etal-2019-deepsentipeer}, as well as that of the Majority Baseline, according to which the prediction is equal to the multiple reviewer recommendation
majority vote. In the case of Individual Peer Review classification, this aggregate prediction is the same for all review samples pertaining to the same paper. We achieve superior performance over the baselines, both when using the meta-review and when fusing all the peer reviews for a specific paper. We note that the meta-review based model is the highest performer as the text is expected to correspond very well to the final recommendation. However, even our regular review fusion model outperforms DeepSentiPeer by a $15.79\%$ absolute increase in accuracy. We hypothesise that this happens due to DeepSentiPeer treating each review (even along with the paper) as a different sample; we use fusion.
Another reason for this improvement is presumably the use of the \texttt{SciBERT} model which was fine-tuned on biomedical and computer science papers.

Table \ref{tab:indiv_baseline} summarises the results for the Individual Peer Review Classification task. From the above experiments, we see that the \texttt{SciBERT} [CSL] approach is superior to sequential GRU.

In Table \ref{tab:top50_autospace}, we present the top 50 important words in the peer reviews for the Individual Peer Review Classification. 


For both meta-reviews and peer reviews, we observe various technical terms (``lda'', ``gans'', ``variational'', ``generative'', ``adversarial'', ``convex'', etc.). While this indicates a positive correlation between some subjects (such as GANs) and the acceptance to ICLR 2017, it is hard to argue about a causal relationship. 
For instance, a possible non-causal relationship could be the following confounding effect: top researchers perform novel research, such as GANs, and top researchers have a higher probability of having their work published.

Furthermore, we observe plenty of terms that seem to directly criticise the quality of the work accomplished in a paper: ``efficiently'', ``confusing'', ``unconvincing'', ``superficial'', ``comprehensive'', ``systematically'', ``carefully'', ``untrue'' and more. Again,
it is easy to make the mistake of assuming a causal relationship between the ``efficiency'' of an algorithm, suggested in a paper, and the acceptance for publication. For instance, depending on the subject of a paper, the efficiency may or may not increase the chances of a publication. To elaborate, for a paper about real-time language interpretation, the efficiency of the algorithm may directly influence the acceptance decision. On the contrary, the efficiency of training a GAN model may be inconsequential, as far as the ICLR reviewers are concerned.

\section{Producing Causal Explanations}
\label{sec:causal}

``Unjustified Causal Interpretation'' is one of the pitfalls of global explanations \cite{molnar2020pitfalls}. Simply put, global explanations detect correlation and correlation does not entail causation. 
In order to learn what leads to a scientific paper being accepted for publication, we need causality. 
For instance, words such as ``efficiency'' and ``novelty'' appear to be important. However, there may not exist a causal relationship between them and the acceptance of a paper. Specifically, more theoretical subjects may demand ``novelty'', and the chances of a theoretical subject being accepted may be greater.  Therefore, the correlation of the novelty and the acceptance of a paper may be an artefact of this confounding effect on the subject of the paper.

Adjusting for every possible confounding and mediating effect is not possible for most machine learning models, partially due to lack of ground truth data. Causality in machine learning is limited to taking several strong assumptions and providing a causal interpretation, for as long as those assumptions hold \cite{molnar2020pitfalls}. When text is involved, even more assumptions must be taken, due to its high dimensionality \cite{CausalBERT}. In Appendix \ref{appendix:assumptions}, we discuss our assumptions.
\pf{ Even when such assumptions are violated, our methodology is meaningful since it allows decorrelating explanations of text classifier from the natural language embedding of the text we want to control. }

\subsection{Background}

In Figure \ref{fig:confoundingpr}, we depict the confounding effect. In this context, $Y$ is the target variable (e.g. the acceptance of a paper for publication), $T$ is the text (e.g. a peer review), and $C$ are the confounding variables (e.g. the subject of a paper). 

The task is to extract a deconfounded lexicon $L$ such that $L(T)$ (the set of words of $L$ that exist in the text $T$) is correlated to the target variable $Y$ but not to the confounders $C$ \citep{pryzant-etal-2018-interpretable}. For our purposes, this
means that the extracted lexicon can offer explanations about the acceptance of a paper for publication regardless of the subject of a paper. 
\citet{pryzant-etal-2018-interpretable} have introduced the concept of \emph{Informativeness Coefficient} $I(L)$:
\begin{align*}
    I(L) = \EX[\mbox{Var}[\EX[Y|L(T), C]] | C]
\end{align*}
where $\mbox{Var}[\EX[Y|L(T), C]]$ is the amount of information in the target variable $Y$ that can be explained both by $L(T)$ and the confounder $C$. 
In order to extract a deconfounded lexicon $L$, we must maximise the Informativeness Coefficient $I(L)$. \pf{The use of $I(L)$ as an evaluation of the quality of causal explanations is motivated by the ability of $I(L)$ to measure the causal effect of the text $T$ on $Y$, under certain assumptions \cite{pryzant-etal-2018-deconfounded}}

To further understand this concept, let us consider a lexicon $L$ of fixed length. $L(T)$ contains either words descriptive of $T \setminus C$ or words descriptive of $T \cap C$. Words related to the confounders (from $T \cap C$), in theory, do not increase the Informativeness Coefficient since the confounder $C$ is treated as a fixed variable that directly affects $Y$. For instance, the word ``gan'' in a peer review may not be as important when the subject of the paper (the confounder) is already taken into consideration. On the contrary, words like ``enjoyable'', which are potentially unrelated to the subject of the paper, can contain more information. Therefore, the fewer the words in the lexicon that are related to the confounders $C$, the more deconfounded a lexicon will be, and $I(L)$ will take larger values.

\begin{figure}[h!]
    \centering
    \includegraphics[scale=.6]{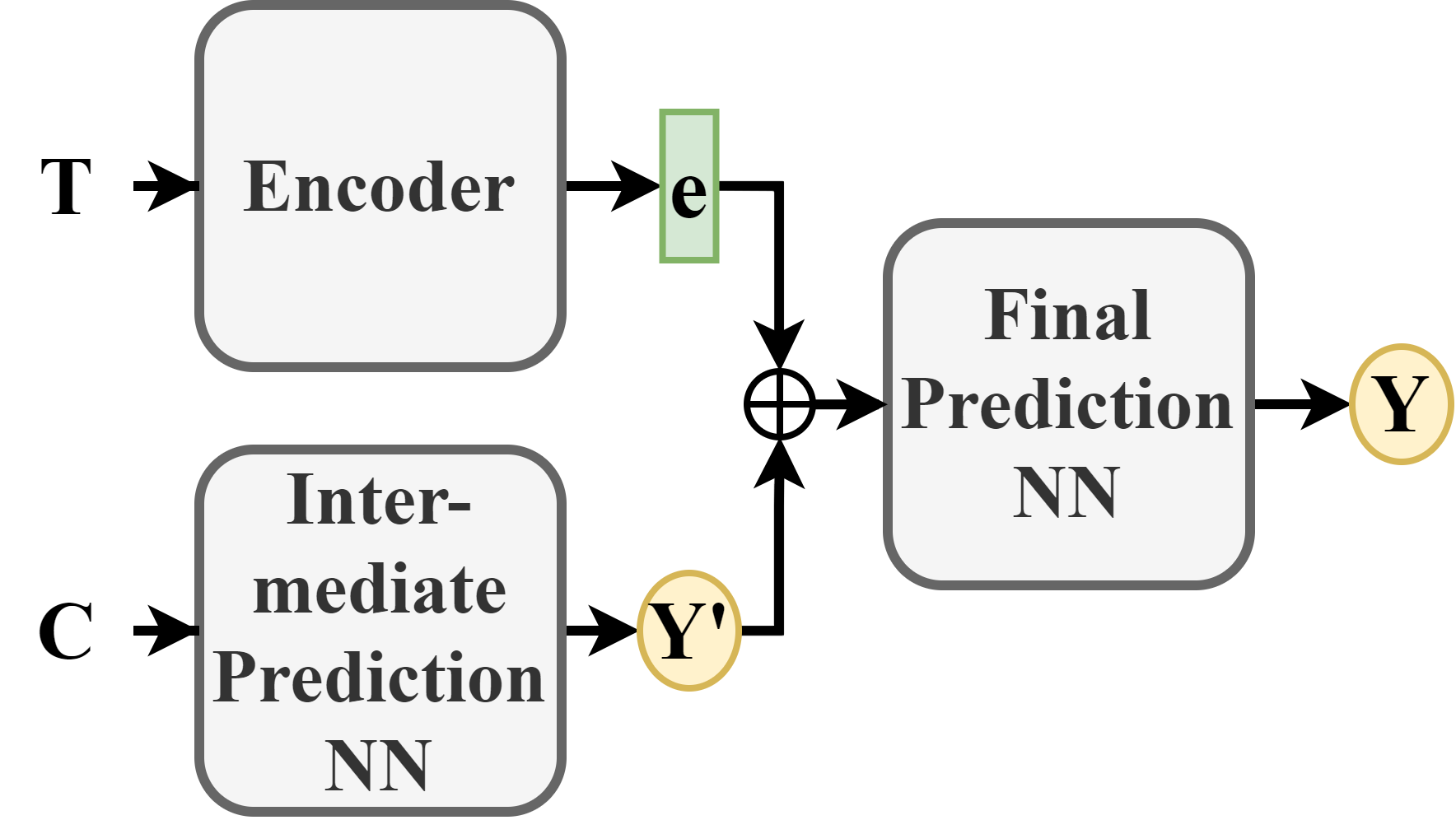}
    \caption{Deep Residualisation Architecture \cite{pryzant-etal-2018-interpretable}. The confounders C  are used for intermediate predictions $Y'$. The treatment T (i.e. the review of a paper) is encoded to a vector $e$, which is concatenated with $Y'$ to produce the final prediction $Y$. }\label{fig:deepresid}
\end{figure}

An ANOVA decomposition is used to write the Informativeness Coefficient of a lexicon $I(L)$ \cite{pryzant-etal-2018-interpretable}:
\begin{align*}
    \begin{split}
        I(L) =  &\EX[(Y - \EX[Y|C])^2] \\
        & -\EX[(Y - \EX[Y|L(T), C])^2]
    \end{split}
\end{align*}
In practice, $I(L)$ can be estimated by fitting a logistic regression on the confounders $C$ to predict the outcome $Y$ and a logistic regression on both $C$ and $L(T)$ (a Bag-of-Word created from the lexicon and the text $T$). Then, for binary classification, the Binary Cross Entropy error is used to measure I(L):
\begin{align*}
    I(L) \simeq BCE_C - BCE_{L(T),C}
\end{align*}
\textbf{Deep Residualisation (DR)} \citep{pryzant-etal-2018-interpretable} is a method for extracting a deconfounded lexicon. A visualisation of the DR model can be seen in Figure \ref{fig:deepresid}. The main idea is to utilise a neural model to predict the target variable $\hat Y'$ using only \textbf{nominal} confounders $C$, and an \textbf{interpretable} model that encodes the text $T$ into an embedding vector $e$. Then, the intermediate prediction $\hat Y'$ is concatenated with the embedding vector $e$ and forwarded through a feed-forward network to produce the final prediction $\hat Y$. The aim of this method is for the encoder to learn to focus on the aspects of the text that are unrelated to the confounders and therefore, maximise the Informativeness Coefficient.


In order to train this model, two loss functions are used. Firstly, the errors from the intermediate prediction $\hat Y'$ (for classification a Binary Cross Entropy function is used) are propagated through the Intermediate Prediction Neural Network. Secondly, the errors from the final prediction $\hat Y$ are propagated through the complete neural network \cite{pryzant-etal-2018-interpretable}.

\subsection{Submission Subject as a Confounder}

Although there may exist multiple confounders to the outcome (e.g., reviewer guidelines, or the paper content itself), we focus specifically on the subject of the paper as a confounder. Such annotation, however, is not part of the available data. Similar to the study performed in \citet{CausalBERT}, we assume that the paper abstract is predictive of its subjects, and opt to utilise it for adjusting for confounding.


A limitation of the work of \cite{pryzant-etal-2018-interpretable} is that they only consider nominal confounders, which they forward through a Multi-layer Perceptron (MLP). We have extended the DR algorithm to extract deconfounded explanations from the reviews while using learnt natural language embeddings as the confounder. Specifically, we are forwarding our confounder (the abstract text) through a SciBERT Language Model to produce an embedding and then, forward that embedding through an MLP, with a sigmoid activation function at the end, to predict the intermediate classification outcome $\hat Y'$. For the task of meta-review classification, $\hat Y'$ aims to predict the final acceptance outcome, and for the task of individual peer review classification,  $\hat Y'$ aims to predict the recommendation of the reviewer. Another assumption has been made here: the embedding method is able to extract information relevant to the subject of the paper. In the Appendix, we present a more detailed discussion into our assumptions.

Firstly, we have extended the \textbf{DR+ATTN} and \textbf{DR+BoW} models \citep{pryzant-etal-2018-interpretable} to allow for treating of natural language as the confounder, using the aforementioned methodology. The DR+ATTN model utilises an RNN-GRU layer with an attention mechanism to encode the text into the vector $e$. Therefore, interpretability is achieved through the attention mechanism. The DR+BoW model, forwards a Bag-of-Words (BoW) representation of the text through a single linear neural layer to produce an one-dimensional vector $e$.
Each feature of our model is an n-gram that may appear on the text of the review. Therefore, we can globally interpret our model by using the weight of an n-gram in the single linear layer as the importance of that word \cite{pryzant-etal-2018-interpretable}.

Secondly, instead of an interpretable model, we utilise a black-box model to encode the text into the vector $e$ and then, use LIME to explain the predictions of our classifier. Therefore, we have developed the \textbf{DR+LIME} variant which can be used to extract a deconfounded lexicon from BERT-based models.


\subsection{Quantitative Evaluation}


\noindent In Table \ref{tab:deconf_measure}, we present metrics for benchmarking the performance of our various lexicons. In order to add more clarity, apart from the Informativeness Coefficient $I(L)$, we include the performance (F1-score) of the different logistic regression models. 

\begin{table*}
\centering
{\fontsize{9.5pt}{9.5pt}\selectfont
\begin{tabular}{cccccc}
\hline
\multirow{3}{*}{\textbf{Model Type}}                                                   & \multirow{3}{*}{\textbf{Model}} &    \multirow{3}{*}{$\boldsymbol{I(L)}$}              & \multicolumn{3}{c}{\textbf{Logistic Regression Macro F1-Score}}                 \\ \cline{4-6} 
                                                                        &                        &          & \textbf{Only Text}  &\textbf{ Only Conf.} & \textbf{Text and Conf.} \\ 
& & &$\boldsymbol{L(T)}$ & $\boldsymbol{C}$&   $\boldsymbol{L(T),C}$                                                         
                                                                        \\\hline
\multirow{2}{*}{\begin{tabular}[]{@{}c@{}}Meta Reviews\end{tabular}} & \multicolumn{1}{l}{GRU}                    & 0.1862          & 0.71      & 0.50             & 0.74                 \\
                                                                        &\multicolumn{1}{l}{DR+ATTN (GRU)}      & \textbf{0.2113} & 0.65      & 0.50             & \textbf{0.75}        \\ \hline
\multirow{6}{*}{\begin{tabular}[c]{@{}c@{}}Individual \\ Peer Reviews\end{tabular}} & \multicolumn{1}{l}{GRU}                    & 0.0252          & 0.60      & 0.55             & 0.60                 \\
                                                                        &\multicolumn{1}{l}{DR+ATTN (GRU)}  & \textbf{0.0907} & 0.59      & 0.55             & \textbf{0.64}                 \\ \cline{2-6} 
                                                                        
                                                                        & \multicolumn{1}{l}{SciBERT}                   & 0.0115          & 0.58      & 0.55             & 0.58                 \\
                                                                        & DR+LIME  (SciBERT)               & \textbf{0.0301} & 0.56      & 0.55             & \textbf{0.60 }                \\ \cline{2-6}
                                                             & \multicolumn{1}{l}{BoW}                    & 0.0039          & 0.43      & 0.55             & 0.55                 \\
                                                                        & \multicolumn{1}{l}{DR+BoW}                & \textbf{0.0128} & 0.43      & 0.56             & \textbf{0.56}                 \\            
                                                                        \hline
\end{tabular}
}
\caption{Performance of the Deconfounded lexicons. All lexicons have a size of 50 words. For each type of classifier, we present the Informativeness Coefficient both of the non-causal lexicon and of the deconfounded lexicon, generated through Deep Residualisation (DR).}
\label{tab:deconf_measure}
\end{table*}

The Informativeness Coefficient is useful for measuring
the degree to which the lexicons are deconfounded. However, the metric by itself is useful for comparing lexicons of \textbf{fixed size} and for \textbf{a specific task}. Therefore, comparing the $I(L)$ of the meta-review with the one from peer reviews is an uneven comparison. The reason for this is that, as we have observed, classifying meta-reviews is a much simpler problem. This leads to lexicons that perform much better and therefore, having higher $I(L)$ values even when words related to the confounders persist in the lexicon.

An important observation is that the deconfounded lexicons are less predictive of the final outcome compared to the non-causal lexicons, when not combined with the Confounder. For instance, in the case of Peer-Review GRU model, the F1-Score of $L(T)$ drops from $60\%$ to $59\%$  with Deep Residualisation.
On the contrary, when the deconfounded lexicons are coupled with the confounders, the logistic regression performs better compared to the non-causal lexicon. For the Peer-Review GRU model, the F1-Score of $L(T),C$ increases from $60\%$ to $64\%$  with Deep Residualisation. The intuition behind this is that the initial lexicon had some words related to the confounders that were important for the final prediction. The DR method does not attend to those words because they are related to the confounders. Instead, they are replaced with words, which may be less predictive of the final outcome, yet uncorrelated to the confounders.

Ultimately, DR+LIME and DR+BoW are less successful than DR+ATTN in generating a deconfounded lexicon for the peer reviews, since the lexicon generated from DR+ATTN achieves superior Informativeness Coefficient. Still, our DR+LIME technique manages to produce a more informative lexicon than DR+BoW.



    
    
    


\subsection{Lexicons Inspection}

 As long as our assumptions hold, we can causally interpret the explanations of our models.
 Since we value the degree with which DR has managed to deconfound our lexicon, the explanations for the peer reviews should be extracted from DR+ATTN.

In Table \ref{tab:top50_autospace}, we can observe a comparison of the top 50 salient words between the causal and non-causal models for the peer reviews.

{
\begin{table*}[h!]
\centering
{\fontsize{8.5pt}{8.5pt}\selectfont
\begin{tabular}{p{0.231\linewidth} p{0.246\linewidth} p{0.231\linewidth} p{0.231\linewidth}}
\hline
\textbf{SciBERT (Non-Causal)}                                                                                                                                                                                                                                                                                                                                                                                                                                           & \textbf{SciBERT (DR+LIME)}                                                                                                                                                                                                                                                                                                                                                                                                                                                                & \textbf{GRU (Non-Causal)}                                                                                                                                                                                                                                                                                                                                                                                                                                         & \textbf{GRU (DR+ATTN)}                                                                                                                                                                                                                                                                                                                                                                                                                                                          \\ \hline
interspeech, zhang, prons, dismissed, third, p6, confusing, resulting, geometry, unconvincing, honestly, not, community, cannot, superficial, readership, big, suggestion, dialogue, revisit, bag, analysed, icml, taken, typo, submitted, energy, nice, spirit, competitive, per, highlighted, multimodality, far, 04562, lack, preprint, dcgan, conduct, word2vec, wen, gan, rejected, start, towards, multiplication, generalisation, auxiliary, parametric, enjoyed & interspeech, dismissed, prons, geometry, p6, submitted, unfair, unconvincing, readership, honestly, not, spirit, confusing, analysed, bag, unable, cannot, rejected, insightful, multiplication, highlighted, enjoyed, misleading, disagree, dialogue,mu\_, wen, lack, nice, multimodality, welcome, conduct, recommender, encourage, dualnets, thanks, cdl, preprint, 04562, enjoy, revisit, community, appreciate, principled, medical, coding, drnn, accompanying, factorization, jmlr & not, attempt, comprehensive, interfere, variational, entirely, systematically, perfect, sure, lda, formalize, tolerant, valid, belief, gan, imposible, adagrad, semantically, aware, clear, carefully, j., and/or, arguably, fully, offer, vs., confident, message, object, receive, probably, multimodality, strictly, directly, twitter, join, e.g., observe, complete, explain, novelty, dualnets, convince, handle, nintendo, theorem, satisfy, ready, demand & identical, sure, premise, carefully, heavy, specify, tease, interfere, necessarily, adagrad, attempt, novelty, fully, repeat, systematically, not, gray, theoretically, anymore, role, belief, recommendation, almost, consistently, primary, subtraction, good, satisfactory, background, write, compute, easy, teach, enough, convince, except, ready, explain, usage, sell, unfortunately, arguably, yet, especially, elegantly, sufficiently, handle, cdl, setup, reception \\ \hline
\end{tabular}
\caption{Top 50 salient words for Individual Peer Review Classification. We present class-agnostic explanations both for some of the non-causal models (Section \ref{sec:data_expl}) and for some DR models (Section \ref{sec:causal}).
}
\label{tab:top50_autospace}
}
\end{table*}
}

In the first place, we can observe that in the top 50 explanations of DR+ATTN, technical words, like ``generative'', ``variational'' and ``lda'' disappear in the explanations of the causal model. This occurrence fits our expectations: the technical words are generally related to the confounder (the subject of a paper) and should not appear in the explanation. 
As far as the BERT-based causal model is concerned, its lexicon still has plenty of technical terms. This is because the DR+LIME method is less successful in removing the confounding effect.
 


 For the meta-reviews, we notice the prevalence of words like ``enjoyed'', ``accepted'', ``mixed'', ``positively'' and ``recommend''. Those words are, for the most part, related with to the opinions of the peer reviewers (e.g. ``The reviewers all \textbf{enjoyed} reading this paper''). Furthermore, a logistic regressor fitted on the average recommendation score of the reviewers is able to predict the final outcome of the submission with an accuracy of $95\%$. All this evidence suggests that the ICLR 2017 committee chair for the vast majority of papers followed the recommendation of the reviewers.

In the case of peer reviews, highly ranked in our explanations, we can observe the word ``novelty''. We can deduce that the Originality of a paper is indeed essential for ICLR 2017.
In the top 50 words we observe words such as ``explain'', ``carefully'', ``systematically'', ``consistently'', ``convince'' and ``sufficiently''. Those words indicate the importance of the Soundness of the scientific approach and how convincingly the claims are presented.
Moreover, in the top 100 words, there words that reflect the Clarity and the quality of writing: ``elegantly'', ``polish'', ``readable'', ``clear''.
We can, also, detect words like ``comprehensive'' and ``extensive'', which reflect the Substance (and completeness) of a paper. Lastly, various words exist that potentially reflect other aspects of a paper, such as Meaningful Comparison (``compare''), Impact (``contribute'') and Appropriateness (``appropriate'').

Therefore, if our assumptions hold, we can claim that certain characteristics of a paper, such as Originality, Soundness and Clarity, lead the reviewers to recommend that a scientific paper is accepted for publication.
 
 \section{Conclusions}
 
Using the peer reviewing process as a use case, 
we have proposed a framework
for interpreting the decisions of text classifiers' causally, by using natural language to adjust for confounding effects. Our methodology succeeds in extracting deconfounded lexicons that exclude technical terms related to the paper subject (i.e., the confounders), a task where established, non-causal global explanation approaches fail. Prior work was limited to either using nominal confounders \cite{Pryzant2017PredictingSF, pryzant-etal-2018-interpretable, pryzant-etal-2018-deconfounded} or binary treatment variables \cite{CausalBERT}. Furthermore, we have extended the Deep Residualisation algorithm from \citet{pryzant-etal-2018-interpretable} to allow for black-box models. 

As is standard in causal inference \cite{molnar2020pitfalls}, the quality of the deconfounded lexicon we generate is limited by certain assumptions. Future work could explore strengthening our assumptions through fine-tuning of the Transformer Language Model representation of the confounder text. Furthermore, fine-tuning the Encoder of DR+LIME may lead to producing more informative lexicons. To avoid overfitting in this fine-tuning process, we recommend a larger dataset than the one used in this paper for peer reviews ($1.3k$ samples). 

\bibliography{anthology,custom}
\bibliographystyle{acl_natbib}

\appendix

\section*{Appendix A: A Discussion on our Assumptions}
\label{appendix:assumptions}

Causal inference is limited to making several assumptions and providing causal guarantees for as long as those assumptions hold \cite{molnar2020pitfalls, pryzant-etal-2018-interpretable, CausalBERT}. We have taken the following assumptions:
\begin{enumerate}
    \item The subject of a paper is sufficient to identify the causal effect, i.e., there is no unobserved confounding. This is a standard causality assumption, albeit a strong one \cite{CausalBERT}. This assumption essentially means that there should not be other external confounders. 
    The first way to attack this assumption is to argue that a reviewer could have biases that affect their decisions. Those biases can act as external confounders. For instance, a famous author may receive preferential treatment from some reviewers. Indeed, in ``single-blind'' peer reviews (as was the case for ICLR 2017) it has been observed that reviewers tend to accept more papers from top universities and companies \cite{Tomkins12708}, compared to ``double-blind'' peer reviews, where the author is anonymous during the review process. However, it is practically impossible to adjust for every potential source of reviewer bias.
    Another way to attack this assumption is by arguing that the writing quality, for instance, could be another confounder. Nevertheless, when a paper lacks clarity, we can assume that the reviewer is going to point out the poor writing quality in the review. If that is the case, the review text accounts for the writing quality and therefore, it is not an external confounder.
    
    \item The abstract of the paper can act as a proxy to its subject. This is the same assumption that \citet{CausalBERT} make. Indeed, an abstract should reflect the main topic of the paper. In that setting, the abstract can be thought of as a ``noisy realisation'' of the subject \cite{CausalBERT}.
    
    \item The embedding method is able to extract information relevant to the subject of the paper. \citet{CausalBERT} suggest that this assumption is justified since we are using BERT embeddings, which achieves state-of-the-art performance on natural language tasks. As they claim: ``this is reasonable in settings where we expect the confounding to be in aspects such as topic, writing quality, or sentiment'' \citep{CausalBERT}. We could further bolster this assumption by fine-tuning our BERT model.

\end{enumerate}

\section*{Appendix B: Addition Information}

Dataset statistics are summarised in Table~\ref{tab:peerread_descr}.

\begin{table}[]
\centering
\begin{tabular}{lcc}
\hline
\textbf{Partition} & \multicolumn{1}{l}{\textbf{\#Papers}} & \multicolumn{1}{l}{\textbf{Accepted / Rejected}} \\ \hline
Training Set       & 349                                   & 139 / 210                                        \\
Validation Set     & 40                                    & 18 / 22                                          \\
Test Set           & 38                                    & 15 / 23                                          \\
Total              & 427                                   & 172 / 255                                        \\ \hline
\end{tabular}
\caption{Description of ICLR 2017 section of PeerRead.}
\label{tab:peerread_descr}
\end{table}

\end{document}